\documentclass[runningheads]{llncs}

\usepackage{eccv}

\usepackage{eccvabbrv}
\usepackage{setspace}

\renewcommand{\arraystretch}{1.18}

\definecolor{oursbg}{RGB}{255, 243, 205}     
\definecolor{weaknum}{RGB}{160, 160, 160}    
\definecolor{headercolor}{RGB}{240, 240, 240}
\usepackage{graphicx}
\usepackage{subcaption}
\usepackage{booktabs}

\usepackage{multirow}

\usepackage[accsupp]{axessibility}  

\usepackage{hyperref}
\usepackage{orcidlink}
\usepackage{bbding}
\usepackage{pifont}

\usepackage{xcolor}
\hypersetup{
    colorlinks=true,        
    urlcolor=magenta,       
}

\newcommand{\ours}{\textbf{\textsc{VidDoS}}\xspace}
\begin{document}

\typeout{TEXTWIDTH=\the\textwidth}
\title{\ours: Universal Denial-of-Service Attack on Video-based Large Language Models} 

\titlerunning{VidDoS: Universal DoS Attack on Video-based LLMs}

\author{Duoxun Tang\inst{1}\orcidlink{0009-0001-0463-8023} \and
Dasen Dai\inst{2}\orcidlink{0009-0000-3227-3140} \and
Jiyao Wang\inst{5}\orcidlink{0000-0002-0743-0121} \and
Xiao Yang\inst{6}\orcidlink{0000-0001-6096-9903} \and \\
Jianyu Wang\inst{1} \and
Siqi Cai\inst{3,4}~\textsuperscript{\Envelope}\orcidlink{0000-0003-3282-9246}}

\authorrunning{Tang et al.}

\institute{Shenzhen International Graduate School, Tsinghua University \and
The Chinese University of Hong Kong, Hong Kong SAR \and Harbin Institute of Technology, Shenzhen \and Shenzhen Loop Area Institute, China \and McGill University \\ \and The Hong Kong University of Science and Technology, Guangzhou
\email{tdx25@mails.tsinghua.edu.cn, 1155211130@link.cuhk.edu.hk}
}

\maketitle
\let\thefootnote\relax\footnotetext{\textsuperscript{\Envelope} Corresponding Author: caisiqi@hit.edu.cn. \\
~~ Code: \url{https://github.com/DAIDASEN/VIDDos}.}

\begin{abstract}
  Video-LLMs are increasingly deployed in safety-critical applications but are vulnerable to Energy-Latency Attacks (ELAs) that exhaust computational resources. Current image-centric methods fail because temporal aggregation mechanisms dilute individual frame perturbations. Additionally, real-time demands make instance-wise optimization impractical for continuous video streams. We introduce \ours, which is the first universal ELA framework tailored for Video-LLMs. Our method leverages universal optimization to create instance-agnostic triggers that require no inference-time gradient calculation. We achieve this through \textit{masked teacher forcing} to steer models toward expensive target sequences,  combined with a \textit{refusal penalty} and \textit{early-termination suppression} to override conciseness priors. Testing across three mainstream Video-LLMs and three video datasets, which include video question answering and autonomous driving scenarios, shows extreme degradation. \ours induces a token expansion of more than 205$\times$ and inflates the inference latency by more than 15$\times$ relative to clean baselines. Simulations of real-time autonomous driving streams further reveal that this induced latency leads to critical safety violations. We urge the community to recognize and mitigate these high-hazard ELA in Video-LLMs.
\end{abstract}

\section{Introduction}
\label{sec:Introduction}

The rapid advancement of Video-LLMs \cite{bai2023qwen,zhang2023video,wang2024qwen2,liu2024improved,li2024llava,maaz2024video} has extended multimodal understanding to dynamic events, accelerating their deployment in safety-critical domains like autonomous driving \cite{xu2024vlm,sima2024drivelm,zhou2024embodied,huang2025vlm}. However, the ever-increasing parameter counts, widespread deployment demands, and substantial energy consumption of large language models render these systems inherently vulnerable to denial-of-service-like threats such as Energy-Latency Attacks (ELAs) \cite{shumailov2021sponge, chen2022nicgslowdown, dong2024engorgio, gao2024denial, gao2024inducing, gao2025recalled, cina2025energy, zhang2025crabs,brachemi2025energy}. By manipulating inputs to force prolonged generation, adversaries can exhaust resources and induce debilitating latency. This threat is particularly acute in real-time applications such as manual takeover \cite{sae2021j3016}, a mechanism increasingly adopted by entities like Wayve, Tesla, and Li Auto. In these high-stakes scenarios, an ELA that forces a verbose response instead of a concise decision can delay reaction times beyond safe limits, directly endangering passenger safety.

While ELAs have been extensively studied in NLP~\cite{shumailov2021sponge}, recent research has begun to extend this threat surface to the visual modality. Specific attacks such as Verbose Images \cite{gao2024inducing} and RECALLED \cite{gao2025recalled} have demonstrated that imperceptible perturbations on static images can manipulate VLMs into generating abnormally long captions. 
However, extending these image-centric strategies to Video-LLMs encounters unique challenges that render naive adaptations ineffective: \ding{182} Video architectures employ aggressive temporal subsampling and pooling. Consequently, perturbations on individual frames are often diluted during feature aggregation, preventing the attack signal from reaching the decoder. \ding{183} Real-time scenarios like autonomous driving demand ultra-low latency. Existing instance-wise optimization methods require expensive gradient calculation for every frame, making them impractical for attacking continuous video streams. \ding{184} Video streams contain dynamic, shifting visual contexts that challenge the robustness of full-frame perturbations. Image-centric attacks typically optimize noise tied to specific static backgrounds, which fails to generalize across varying temporal frames. This exposes the absence of a spatially concentrated, content-agnostic optimization strategy (e.g., universal patches) capable of anchoring the model's attention independent of the changing video content. Consequently, current red-teaming protocols exhibit a significant attack capability gap when targeting Video-LLMs.

To bridge this gap, we introduce \ours, the first universal Energy-Latency Attack framework specifically tailored for Video-LLMs. Unlike prior instance-specific optimizations that are often diluted by temporal aggregation, \ours leverages a spatially concentrated universal patch optimized over a surrogate dataset. This strategy bypasses the structural "low-pass filter" effect of video encoders by anchoring cross-modal attention to a localized semantic anomaly. Our methodology fundamentally shifts the attack paradigm from pixel-wise noise to trajectory steering, employing two novel mechanisms: (1) a Masked Teacher Forcing objective that aligns the model’s predictive distribution with a weighted, computationally expensive "sponge" sequence, and (2) an Early-Termination Suppression penalty that explicitly discourages the emission of 'End-of-Sequence' (EOS) tokens or concise responses (e.g., "Yes/No"). By injecting these content-agnostic triggers into the visual stream, we covertly hijack the decoder's trajectory, forcing it into a pathologically long generation regime that induces severe inference delays without altering the textual prompt. \ours achieves a 'train-once, deploy-anywhere' efficiency: once the universal trigger is optimized on a surrogate dataset, it can be instantly applied to any unseen video stream in real-time without requiring any inference-time gradient computations. Moreover, \ours is resistant to stochastic noise introduced by higher temperatures.

Our contributions are summarized as follows:
\begin{enumerate}
\item We propose \ours, the first universal ELA for Video-LLMs, leveraging a spatially concentrated patch to resist aggressive temporal subsampling, pooling, and stochastic noise from higher decoding temperatures.

\item We introduce a specialized optimization framework combining \textit{masked teacher forcing}, \textit{refusal penalty} and \textit{early-termination suppression} to override fine-tuned models’ conciseness priors and enable unbounded generation.

\item Extensive empirical evaluation on three actual morden Video-LLMs and three datasets demonstrates state-of-the-art attack potency and robust generalization across diverse video scenarios.
\end{enumerate}

\section{Related Work}
\label{sec:related_work}

\subsection{Video Understanding Models}

The rapid evolution of Multimodal Large Language Models (MLLMs) \cite{fang2024mmbench, huang2024vtimellm, song2024moviechat, yuan2025videorefer, shu2025audio, yang2026model} has bridged the gap between static vision and temporal dynamics. Mainstream Video-LLMs, including the LLaVA series~\cite{liu2024improved,li2024llava}, the Qwen-VL family~\cite{bai2023qwen,wang2024qwen2}, Video-LLaVA~\cite{lin2024video}, and TimeChat~\cite{ren2024timechat}, generally adopt a sophisticated two-stage processing pipeline. Formally, let $\mathcal{I}_{vid} \in \mathbb{R}^{F \times H \times W \times 3}$ represent the raw input video frames and $\mathcal{I}_{txt}$ denote the textual instruction. A visual backbone $\Omega$ first extracts frame-level features, which are subsequently compressed by an alignment adapter $\Pi$, such as the Q-Former~\cite{li2023blip} in Video-LLaMA~\cite{zhang2023video} or spatio-temporal pooling in Video-ChatGPT~\cite{maaz2024video}, into a dense sequence of visual embeddings $\mathbf{E}_{vis} = \Pi(\Omega(\mathcal{I}_{vid})) \in \mathbb{R}^{L \times D}$, where $L$ denotes the reduced token count essential for efficiency. These visual tokens are concatenated with the instruction embeddings $\mathbf{E}_{txt}$ to construct a unified multimodal context $\mathbf{X} = [\mathbf{E}_{vis}; \mathbf{E}_{txt}]$. The LLM decoder then autoregressively predicts the response sequence $y$ by maximizing the likelihood $p(y_t \mid y_{<t}, \mathbf{X})$. In the safety-critical domain of autonomous driving, this paradigm is specifically adapted to process vehicle-centric data streams, as seen in models like RoboTron-Drive~\cite{huang2024drivemm}. This work introduces latency-oriented resource exhaustion attacks on Video-LLMs and highlights safety risks of large models in autonomous driving.

\subsection{Energy-Latency Attacks}
Energy-Latency Attacks (ELAs) exploit worst-case computational behaviors of deep learning models to degrade service availability, presenting a threat analogous to Denial-of-Service (DoS) attacks in networked systems~\cite{pelechrinis2010denial, somani2016ddos, wang2023energy, kaur2025comprehensive}. 
This line of research originated in textual and unimodal settings. Sponge Examples~\cite{shumailov2021sponge} first demonstrated that inputs optimized to maximize activation norms can trigger energy-intensive inference paths on hardware accelerators. Subsequent work NICGSlowDown~\cite{chen2022nicgslowdown} shifted focus to sequence generation models, crafting perturbed images that suppress End-of-Sequence (EOS) token probabilities to force excessive decoder iterations. More recent text-centric~\cite{li2025loopllm} methods employ adversarial prompting techniques: Engorgio~\cite{dong2024engorgio} and DoS poisoning~\cite{gao2024denial} construct semantically coherent yet resource-intensive queries, while Crabs~\cite{zhang2025crabs} leverages auto-generation under black-box constraints to amplify latency by over 250$\times$. 
The threat surface has recently expanded to vision-language models~\cite{muller2024impact}. Verbose Images~\cite{gao2024inducing} and RECALLED~\cite{gao2025recalled} reveal that imperceptible visual perturbations can induce "babbling" behaviors in VLMs, forcing unbounded text generation and substantially inflating inference latency and GPU resource consumption.

However, a critical security gap remains in the temporal modality, as direct adaptation of these image-centric strategies to Video-LLMs is rendered ineffective by three inherent architectural constraints. 
First, video models utilize robust \textbf{temporal aggregation} mechanisms that tend to filter out frame-specific perturbations as incoherent noise during feature compression, preventing the attack signal from propagating to the decoder. 
Second, the \textbf{efficiency demands} of real-time systems (e.g., autonomous driving) preclude the use of expensive instance-wise gradient optimization required by prior methods, necessitating a universal, inference-free attack vector. 
Third, the \textbf{dynamic visual context} of video streams challenges the robustness of static spatial noise, which fails to anchor the model's attention across shifting temporal frames. 
To date, no framework has effectively addressed these spatio-temporal challenges to evaluate the latency-oriented availability robustness of Video-LLMs.

\section{Methodology}
\label{sec:methodology}

\begin{figure}[!ht]
    \centering
    \includegraphics[width=0.96\linewidth]{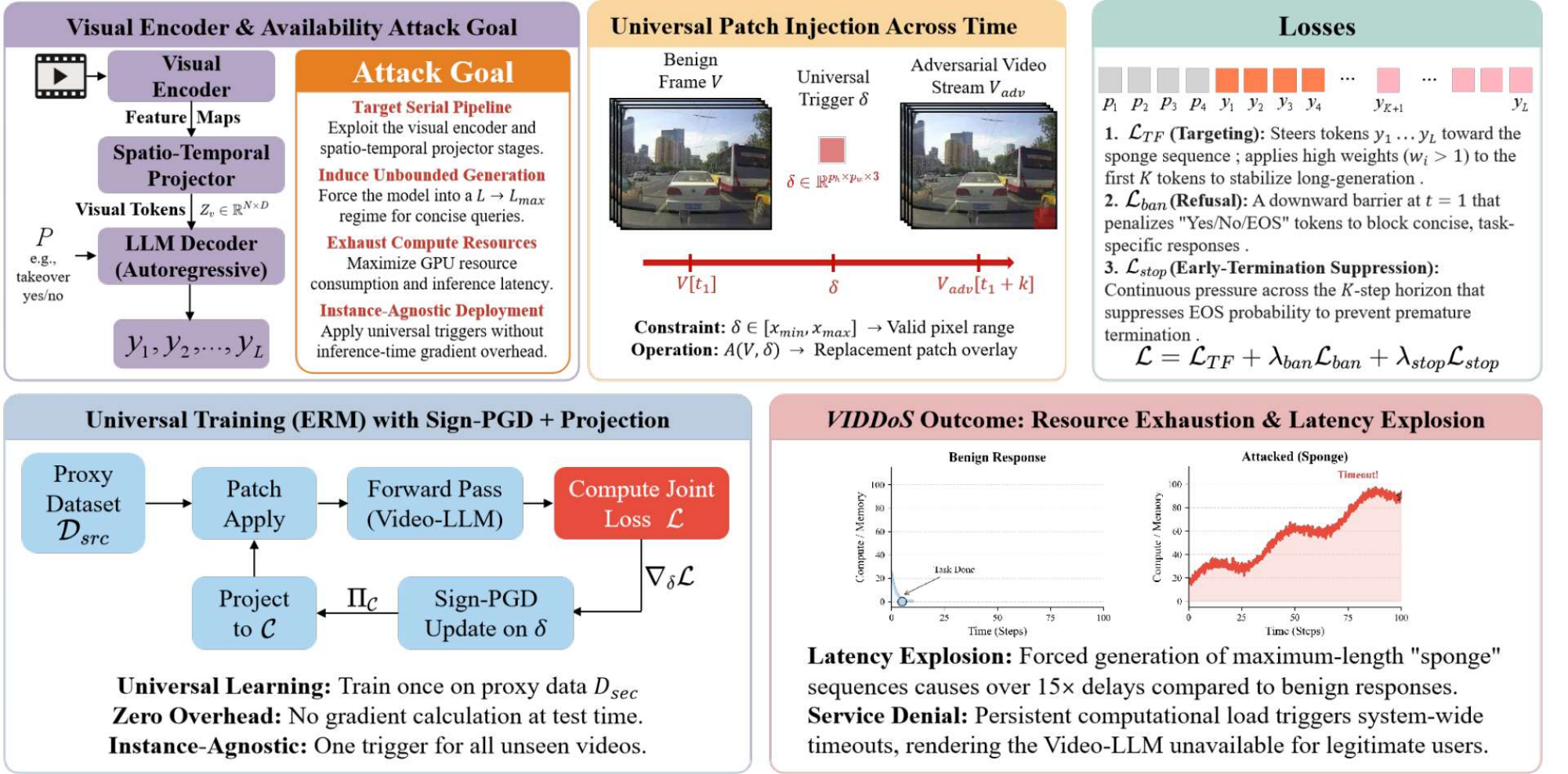}
    \caption{Overview of \ours. The framework involves: (1) Defining the attack goal; (2) Temporal injection of a universal patch; (3) Steering decoder trajectory via joint losses (Masked Teacher Forcing, Refusal Penalty, and ETS); (4) Offline patch optimization via Sign-PGD; and (5) Real-time latency induction.}
    \label{fig:pipline}
\end{figure}

\subsection{Preliminaries and Threat Model}

\subsubsection{Problem Definition.}
Let $\mathcal{M}_\theta$ denote a Video-LLM parameterized by $\theta$. The model processes a continuous video input $\mathbf{V} \in \mathbb{R}^{T \times H \times W \times 3}$ comprising $T$ frames (with spatial resolution $H \times W$ and 3 color channels), alongside a textual prompt $\mathbf{P}$. During the inference forward pass, a visual encoder first maps $\mathbf{V}$ to dense feature representations. These representations are subsequently compressed via a spatio-temporal projector (e.g., pooling mechanisms or Querying Transformers~\cite{li2023blip}) into a distinct sequence of visual embeddings $\mathbf{Z}_v \in \mathbb{R}^{N \times D}$, where $N$ is the number of visual tokens and $D$ is the embedding dimension. Conditioned on the concatenated visual and prompt embeddings $[\mathbf{Z}_v; \mathbf{Z}_p]$, the autoregressive LLM decoder computes the conditional probability distribution to generate a response token sequence $\mathbf{Y} = \{y_1, y_2, \dots, y_L\}$.

Our objective is to construct a \emph{universal adversarial trigger} $\boldsymbol{\delta}$ that is agnostic to specific video content. For any benign video $\mathbf{V}$ sampled from a natural data distribution $\mathcal{D}$, we construct an adversarial video $\mathbf{V}_{\text{adv}} = \mathcal{A}(\mathbf{V}, \boldsymbol{\delta})$, where $\mathcal{A}$ denotes a predefined injection mapping (e.g., spatial patch overlay or additive noise). To ensure the perturbation remains visually imperceptible or inconspicuous to human observers, its magnitude is strictly bounded by an $\ell_\infty$-norm constraint $\|\mathbf{V}_{\text{adv}} - \mathbf{V}\|_\infty \le \epsilon$.

\subsubsection{Threat Model.}
We formulate a Denial-of-Service (DoS) availability attack targeting Video-LLMs deployed in practical streaming contexts (e.g., autonomous driving visual question answering). The adversary aims to artificially inflate the inference latency and exhaust GPU computational resources $\mathcal{C}$ by coercing the autoregressive decoder into an unbounded generation regime ($L \to L_{\max}$) under normally benign, closed-ended queries (e.g., ``Does this scenario require a manual takeover?''). We consider a \textit{white-box} threat model where the adversary possesses full access to the target model's parameters $\theta$ to compute exact gradients. The core requirement for the adversary is to learn a single, unified trigger $\boldsymbol{\delta}$ on a proxy set of public videos $\mathcal{D}_{\text{src}}$, which can subsequently be deployed to reliably attack the victim model on unseen video streams without incurring any instance-specific optimization overhead.

\subsection{Design of the Proposed Attack}

\subsubsection{Motivation.}
Extending availability attacks to Video-LLMs presents unique challenges. Unlike static vision-language models, Video-LLMs utilize temporal subsampling and aggregation that act as structural low-pass filters, diluting diffuse pixel-wise noise before it reaches the decoder. Furthermore, models fine-tuned for discriminative tasks possess a strong inductive prior for conciseness (e.g., "Yes/No"), which naive perturbations fail to override across dynamic contexts. To bypass this temporal dilution, we anchor cross-modal attention using a spatially concentrated universal patch and explicitly steer the decoder’s predictive trajectory toward pathological, cyclical outputs. Figure~\ref{fig:pipline} shows the pipline of \ours.

\subsubsection{Masked Teacher Forcing.}
Let $\Phi(\cdot)$ represent the model's visual pre-processing pipeline mapping the raw video to a normalized tensor $\tilde{\mathbf{x}}$. We construct a structured input sequence $\mathbf{z} = [\mathbf{z}_{\text{prompt}}; \mathbf{z}_{\text{target}}]$ by concatenating the system query with a repetitive target string $\mathbf{y}^\star$ (the semantic ``sponge''). 

To optimize the input towards producing $\mathbf{y}^\star$ while preserving the integrity of the conditional prompt, we employ a \emph{masked teacher forcing} strategy. We compute a weighted token-level cross-entropy loss exclusively on the target assistant tokens. With $L_p = |\mathbf{z}_{\text{prompt}}|$ denoting the boundary index, the targeted generation loss is:
\begin{equation}
\mathcal{L}_{\text{TF}}(\tilde{\mathbf{x}};\mathbf{y}^\star)
=
\frac{1}{\sum w_i} \sum_{i=L_p+1}^{|\mathbf{z}|}
w_i \cdot
\mathrm{CE}\!\left(p_\theta(z_i \mid z_{<i}, \tilde{\mathbf{x}}), z_i\right),
\label{eq:tf_loss}
\end{equation}
where $w_i$ is a positional scaling factor. Crucially, we set $w_i > 1$ for the first $K$ tokens of the target sequence and $w_i = 1$ elsewhere. This asymmetric weighting explicitly anchors the decoder's deviation during the critical early steps to stabilize its entry into the long-generation regime.

\subsubsection{Refusal Penalty and Early-Termination Suppression.}
Because practical Video-LLM applications frequently anticipate concise responses, the decoder exerts a strong bias toward early termination via task-specific short answers (e.g., ``Yes'' or ``No'') or the immediate emission of the End-Of-Sequence (EOS) token. To dismantle this prior, we introduce two auxiliary penalties operating within a shared intervention horizon $K$.

First, to prevent the model from outputting short task-relevant answers or halting at the very beginning, we define a banned vocabulary subset $\mathcal{B}_{\text{ban}}$ (containing ``Yes'', ``No'', and the EOS token). We penalize its cumulative probability mass specifically at the first generation step ($k=1$):
\begin{equation}
\mathcal{L}_{\text{ban}}(\tilde{\mathbf{x}})
=
\sum_{b\in\mathcal{B}_{\text{ban}}}
p_\theta(b \mid z_{<L_p+1}, \tilde{\mathbf{x}}).
\label{eq:ban_loss}
\end{equation}

Second, to ensure the generation does not halt prematurely after bypassing the initial step, we apply an aggressive suppression to the EOS emission probability, averaged over the same prefix horizon $K$:
\begin{equation}
\mathcal{L}_{\text{stop}}(\tilde{\mathbf{x}})
=
\frac{1}{K} \sum_{k=1}^{K}
-\log \Big( 1 - p_\theta(\text{EOS} \mid z_{<L_p+k}, \tilde{\mathbf{x}}) \Big).
\label{eq:stop_loss}
\end{equation}

The joint objective for a single video instance is thus framed as a tripartite optimization:
\begin{equation}
\min_{\tilde{\mathbf{x}}^{\mathrm{adv}}\in \mathcal{S}(\tilde{\mathbf{x}})}
\;\;
\mathcal{L}_{\text{TF}}(\tilde{\mathbf{x}}^{\mathrm{adv}};\mathbf{y}^\star)
+
\lambda_{\text{ban}} \mathcal{L}_{\text{ban}}(\tilde{\mathbf{x}}^{\mathrm{adv}})
+
\lambda_{\text{stop}} \mathcal{L}_{\text{stop}}(\tilde{\mathbf{x}}^{\mathrm{adv}}),
\label{eq:overall_obj}
\end{equation}
where $\mathcal{S}(\tilde{\mathbf{x}})$ enforces the perturbation constraints within the normalized pixel space, and $\lambda_{\text{ban}}, \lambda_{\text{stop}}$ act as hyperparameters balancing the generation steering and termination suppression.

\subsubsection{Universal Optimization via Patch Training.}
To facilitate a zero-overhead attack during inference, we transition the optimization paradigm to learn a \emph{universal} adversarial trigger based on empirical risk minimization over a training distribution. 

\textbf{Patch vs. Point-wise perturbation.} We consider two primary parameterizations for the universal trigger: a full-frame point-wise additive perturbation and a spatially concentrated replacement patch. While pixel-wise perturbations (Point Training) manipulate the entire frame, they are highly susceptible to the low-pass filtering effects of spatio-temporal pooling and tokenization inherent in Video-LLMs. In contrast, a localized replacement patch (patch training) creates a concentrated, high-magnitude semantic anomaly. This dense injection effectively hijacks the cross-modal attention mechanism, rendering it significantly more resilient to the model's internal feature compression. Consequently, we employ the patch parameterization as our primary trigger, while maintaining the full-frame point perturbation as a baseline variant for empirical comparison in subsequent sections.

\textbf{Spatial configuration.} We parameterize the universal trigger as a spatially localized, learnable replacement patch $\boldsymbol{\delta}\in\mathbb{R}^{p_h\times p_w\times 3}$ with height $p_h$ and width $p_w$. Given a spatial coordinate $(u,v)$, the identical patch is symmetrically injected across the temporal dimension $t = 1, \dots, T$:
\begin{equation}
\tilde{\mathbf{x}}^{\mathrm{adv}}_{t}[u{:}u{+}p_h,\;v{:}v{+}p_w]
\leftarrow
\boldsymbol{\delta}.
\label{eq:patch_apply}
\end{equation}
Crucially, the coordinate $(u,v)$ is typically fixed to a peripheral region (e.g., the bottom-right corner) rather than the center. The rationale is to prevent the patch from occluding the primary semantic objects of the driving scene (e.g., front vehicles, traffic lights), thereby ensuring the visual input remains structurally benign to the task prompt while maintaining human imperceptibility. Alternatively, $(u,v)$ can be randomized during training to yield a position-invariant trigger.

The universal optimization is subsequently cast as minimizing the expected loss over the source dataset $\mathcal{D}_{\mathrm{src}}$:
\begin{equation}
\min_{\boldsymbol{\delta}\in \mathcal{C}}
\;\;
\mathbb{E}_{\mathbf{x}\sim \mathcal{D}_{\mathrm{src}}}
\Big[
\mathcal{L}_{\text{TF}}(\tilde{\mathbf{x}}^{\mathrm{adv}};\mathbf{y}^\star)
+
\lambda \,\mathcal{L}_{\text{stop}}(\tilde{\mathbf{x}}^{\mathrm{adv}})
\Big],
\label{eq:universal_obj}
\end{equation}
where $\mathcal{C}$ strictly enforces the valid value dynamic ranges corresponding to the processor's input space constraints. During training, $\boldsymbol{\delta}$ is iteratively updated using sign-based PGD~\cite{madry2017towards} steps. After each update, the patch is strictly projected back onto the feasible set $\mathcal{C}$ to maintain boundary validity while maximizing the computational degradation inflicted upon the decoder.

\section{Experiment}
\label{sec:Experiment}

\subsection{Setup}

\subsubsection{Video datasets.}
To comprehensively evaluate the effectiveness and generalizability of our proposed \ours, we select three diverse video datasets. 
\textbf{BDDX}~\cite{BDDX} and \textbf{D$^2$-City}~\cite{che2019d} provide complex, high-dynamic autonomous driving and diverse traffic dashcam scenarios, which are critical for evaluating the safety and reliability of multimodal systems in real-world critical environments. 
\textbf{VideoSimpleQA}~\cite{videosimpleqa} is utilized as a complementary benchmark for general video question answering. To standardize the evaluation and eliminate the variance of open-ended response lengths, we reformat its original QA pairs into a binary verification task (Yes/No). Specifically, we synthesize affirmative statements from the ground-truth answers (expecting a "Yes") and randomly introduce negated or mismatched statements (expecting a "No"). This ensures that our evaluation spans both specialized high-stakes domains and rigorously controlled general factuality tasks.

\subsubsection{Victim models.}
We target three state-of-the-art open-source Video-LLMs and multimodal foundational models that are widely adopted in the community. Specifically, we evaluate \textbf{LLaVA-NeXT-Video-7B} \cite{zhang2024llavanextvideo}, \textbf{Qwen3-VL-4B-Instruct} \cite{yang2025qwen3}, and \textbf{Video-LLaVA-7B-hf} \cite{lin2024video}. These models represent a diverse set of visual encoders, spatio-temporal projection architectures, and underlying LLM reasoning engines, providing a robust testbed to demonstrate the universal threat and cross-architecture transferability of our attack.

\subsubsection{Competitors.}
We compare our \ours against a clean baseline (No Attack) and three adversarial competitors:
(1) \textbf{Random Noise}: A fundamental baseline applying $\ell_\infty$-bounded random noise to the video frames.
(2) \textbf{Verbose Images}~\cite{gao2024inducing}: An attack originally designed to induce high energy-latency in 2D Vision-Language Models. We adapt it to the video domain by applying the perturbations across the sampled frames.
(3) \textbf{NICGSlowDown}~\cite{chen2022nicgslowdown}: An efficiency degradation attack initially proposed for neural image captioning models, which we similarly adapt to our multi-frame video QA setting to serve as a strong cross-domain baseline.

\subsubsection{Evaluation metrics.}
To quantify the availability impact and the severity of the sponge effect, we run greedy decoding with a fixed generation limit (\texttt{max\_new\_tokens} $= 512$). We measure the attack's effectiveness primarily across two dimensions:
(i) \textbf{Generation Length}: We report the absolute number of generated tokens under attack (\textbf{Adv Tokens}) and calculate the \textbf{Token Ratio} ($\uparrow$) relative to the clean baseline to demonstrate the inflation scale.
(ii) \textbf{End-to-End Runtime}: We measure the total generation latency in seconds (\textbf{Adv Lat.}) and report the absolute latency \textbf{Overhead} ($\uparrow$) induced by the attack. 
\begin{table*}[t]
\centering
\caption{
\textbf{Quantitative comparison of attack methods.} Best results are in \textbf{bold}.
}
\label{tab:main_results_baselines}
\resizebox{\textwidth}{!}{
\begin{tabular}{@{}llcc l cccc@{}}
\toprule
\multirow{2}{*}{\textbf{Dataset}} & \multirow{2}{*}{\textbf{Victim Model}} & \multicolumn{2}{c}{\textbf{Clean (No Attack)}} & \multirow{2}{*}{\textbf{Attack Method}} & \multicolumn{4}{c}{\textbf{Adversarial Performance}} \\ \cmidrule(lr){3-4} \cmidrule(l){6-9}
 &  & Tokens & Latency (s) &  & Adv Tokens & Ratio ($\uparrow$) & Adv Lat. (s) & Overhead ($\uparrow$) \\ \midrule

\multirow{12}{*}{\textbf{BDDX}}
 & \multirow{4}{*}{LLaVA-NeXT-Video-7B} & \multirow{4}{*}{28.40} & \multirow{4}{*}{1.12}
 & Random Noise & 31.35 & 5.15 & 1.22 & -0.05 \\
 & & & & Verbose Images & 31.00 & 5.03 & 1.22 & -0.05 \\
 & & & & NICGSlowDown & 32.50 & 1.74 & 1.19 & 0.07 \\
 & & & & \textbf{\ours (Ours)} & \textbf{333.50} & \textbf{67.13} & \textbf{9.38} & \textbf{8.25} \\ \cmidrule(l){2-9}

 & \multirow{4}{*}{Qwen3-VL-4B-Instruct} & \multirow{4}{*}{2.00} & \multirow{4}{*}{0.16}
 & Random Noise & 2.00 & 1.00 & 0.15 & 0.00 \\
 & & & & Verbose Images & 2.40 & 1.20 & 0.16 & 0.01 \\
 & & & & NICGSlowDown & 5.20 & 2.60 & 0.20 & 0.04 \\
 & & & & \textbf{\ours (Ours)} & \textbf{394.60} & \textbf{197.30} & \textbf{15.57} & \textbf{15.41} \\ \cmidrule(l){2-9}

 & \multirow{4}{*}{Video-LLaVA-7B-hf} & \multirow{4}{*}{2.00} & \multirow{4}{*}{0.38}
 & Random Noise & 2.00 & 1.00 & 0.31 & 0.02 \\
 & & & & Verbose Images & 1.95 & 0.98 & 0.32 & 0.02 \\
 & & & & NICGSlowDown & 3.95 & 1.98 & 0.41 & 0.03 \\
 & & & & \textbf{\ours (Ours)} & \textbf{411.45} & \textbf{205.73} & \textbf{11.47} & \textbf{11.09} \\ \midrule

\multirow{12}{*}{\textbf{VideoSimpleQA}}
 & \multirow{4}{*}{LLaVA-NeXT-Video-7B} & \multirow{4}{*}{89.40} & \multirow{4}{*}{2.72}
 & Random Noise & 96.25 & 1.31 & 2.98 & 0.18 \\
 & & & & Verbose Images & 96.25 & 1.32 & 2.99 & 0.18 \\
 & & & & NICGSlowDown & 94.15 & 1.08 & 2.78 & 0.07 \\
 & & & & \textbf{\ours (Ours)} & \textbf{352.15} & \textbf{6.52} & \textbf{9.69} & \textbf{6.97} \\ \cmidrule(l){2-9}

 & \multirow{4}{*}{Qwen3-VL-4B-Instruct} & \multirow{4}{*}{28.95} & \multirow{4}{*}{1.22}
 & Random Noise & 86.80 & 1.16 & 3.56 & -0.09 \\
 & & & & Verbose Images & 86.55 & 1.10 & 3.56 & -0.09 \\
 & & & & NICGSlowDown & 31.30 & 2.18 & 1.26 & 0.04 \\
 & & & & \textbf{\ours (Ours)} & \textbf{169.55} & \textbf{30.61} & \textbf{6.86} & \textbf{5.64} \\ \cmidrule(l){2-9}

 & \multirow{4}{*}{Video-LLaVA-7B-hf} & \multirow{4}{*}{30.65} & \multirow{4}{*}{1.05}
 & Random Noise & 28.30 & 0.96 & 1.01 & -0.04 \\
 & & & & Verbose Images & 28.50 & 0.97 & 1.01 & -0.05 \\
 & & & & NICGSlowDown & 32.40 & 1.04 & 1.13 & 0.08 \\
 & & & & \textbf{\ours (Ours)} & \textbf{462.55} & \textbf{17.92} & \textbf{12.67} & \textbf{11.62} \\ \midrule

\multirow{12}{*}{\textbf{D$^2$-City}}
 & \multirow{4}{*}{LLaVA-NeXT-Video-7B} & \multirow{4}{*}{3.00} & \multirow{4}{*}{0.70}
 & Random Noise & 3.00 & 1.00 & 0.67 & -0.03 \\
 & & & & Verbose Images & 2.60 & 0.87 & 0.67 & -0.03 \\
 & & & & NICGSlowDown & 5.95 & 1.98 & 0.73 & 0.08 \\
 & & & & \textbf{\ours (Ours)} & \textbf{163.80} & \textbf{54.60} & \textbf{4.90} & \textbf{4.25} \\ \cmidrule(l){2-9}

 & \multirow{4}{*}{Qwen3-VL-4B-Instruct} & \multirow{4}{*}{2.00} & \multirow{4}{*}{0.73}
 & Random Noise & 2.00 & 1.00 & 0.72 & -0.01 \\
 & & & & Verbose Images & 2.15 & 1.08 & 0.72 & 0.00 \\
 & & & & NICGSlowDown & 6.90 & 3.45 & 0.81 & 0.08 \\
 & & & & \textbf{\ours (Ours)} & \textbf{254.80} & \textbf{127.40} & \textbf{10.66} & \textbf{9.93} \\ \cmidrule(l){2-9}

 & \multirow{4}{*}{Video-LLaVA-7B-hf} & \multirow{4}{*}{2.00} & \multirow{4}{*}{0.38}
 & Random Noise & 2.00 & 1.00 & 0.40 & 0.03 \\
 & & & & Verbose Images & 1.50 & 0.75 & 0.40 & 0.03 \\
 & & & & NICGSlowDown & 2.00 & 1.00 & 0.38 & 0.00 \\
 & & & & \textbf{\ours (Ours)} & \textbf{233.60} & \textbf{116.80} & \textbf{6.64} & \textbf{6.27} \\ \bottomrule
\end{tabular}
}
\end{table*}

\subsection{Attack Results}
\subsubsection{In-domain Results.}
Table 1 presents the quantitative evaluation of \ours against state-of-the-art Video-LLMs. The empirical results yield several key observations:

\textbf{Baselines exhibit negligible impact on Video-LLMs}. Existing image-centric attacks, such as Verbose Images and NICGSlowDown, fail to generalize to the video modality, often yielding token ratios near $1.0\times$ and negligible or even negative latency overhead. For instance, on the BDDX dataset with Qwen3-VL, NICGSlowDown only achieves a $2.6\times$ token ratio. This confirms our theoretical motivation that diffuse, frame-specific perturbations are effectively filtered out by the aggressive temporal aggregation and structural low-pass filtering inherent in Video-LLM architectures.

\textbf{\ours demonstrates extreme attack potency.} In contrast, our proposed method consistently induces massive computational exhaustion across all tested models and datasets. On the BDDX dataset, \ours steers the Qwen3-VL model into a pathologically verbose state, achieving a 197.3$\times$ token expansion and inflating inference latency by 15.5$\times$ relative to the clean baseline. Similarly, on the $D^2$-City dataset, the attack maintains high effectiveness with an overhead of 9.9$\times$, highlighting its ability to persistently override the model's conciseness priors even in high-dynamic autonomous driving contexts.

\textbf{Methodological effectiveness across diverse tasks.} The success of VidDoS is particularly evident in the VideoSimpleQA benchmark. For the Video-LLaVA model, our attack forces the generation of 462.55 tokens (a 17.9$\times$ ratio) from a query that normally requires only 30.65 tokens, resulting in an overhead of 11.6$\times$. This extreme "Sponge" effect validates our \textit{masked teacher forcing} and \textit{refusal penalty} mechanisms, which successfully hijack the decoder's predictive trajectory. By anchoring cross-modal attention via a spatially concentrated patch, \ours ensures that the adversarial signal survives feature compression, forcing the LLM backbone into an expensive, near-unbounded generation regime across both general and specialized domains.

\subsubsection{Cross-datasets Transferability.}
We present the results of the cross-domain evaluation in Figure~\ref{fig:distribution} (a). To rigorously assess the transferability of our spatial-aware sponge attack, we conducted a $3 \times 3$ cross-dataset evaluation matrix using LLaVA-NeXT-Video-7B across BDDX, DD, and VideoSimpleQA. 

As illustrated, our attack demonstrates exceptional in-domain efficacy, easily reaching the maximum generation limit (512 tokens) on both BDDX and VideoSimpleQA. More importantly, the attack exhibits strong cross-domain transferability within semantically similar domains. Notably, the universal adversarial patch trained on the BDDX dataset effectively transfers to the DD dataset and yields an average of 486.55 generated tokens. This high transferability between driving datasets, despite their variance in original aspect ratios and resolutions, highlights a crucial architectural vulnerability in models utilizing 5D spatio-temporal tensors (e.g., Batch, Time, Channel, Height, Width). Unlike dynamic 1D sequence flattening, which often suffers from spatial coordinate misalignment, the rigid grid structure inherently anchors the adversarial patch to the targeted physical receptive field (e.g., the bottom-right corner), making the adversarial features highly resilient to cross-dataset resolution shifts.

However, a noticeable performance drop occurs when transferring patches between drastically different semantic domains (e.g., BDDX to VideoSimpleQA yielding 63.00 tokens). This indicates that while the spatial anchoring remains intact, the adversarial trigger overfits to the source domain's visual priors (such as dashboards or road textures), highlighting the natural boundary of domain-specific semantic alignment.

\subsection{Safety Analysis in Autonomous Driving}

To mirror real-world deployment, our setup evaluates a real-time onboard inference pipeline using a streaming sliding-window mechanism. Unlike offline benchmarks limited to isolated video clips, this approach processes continuous driving streams under realistic temporal constraints:

\begin{figure}
    \centering
    \includegraphics[width=0.75\linewidth]{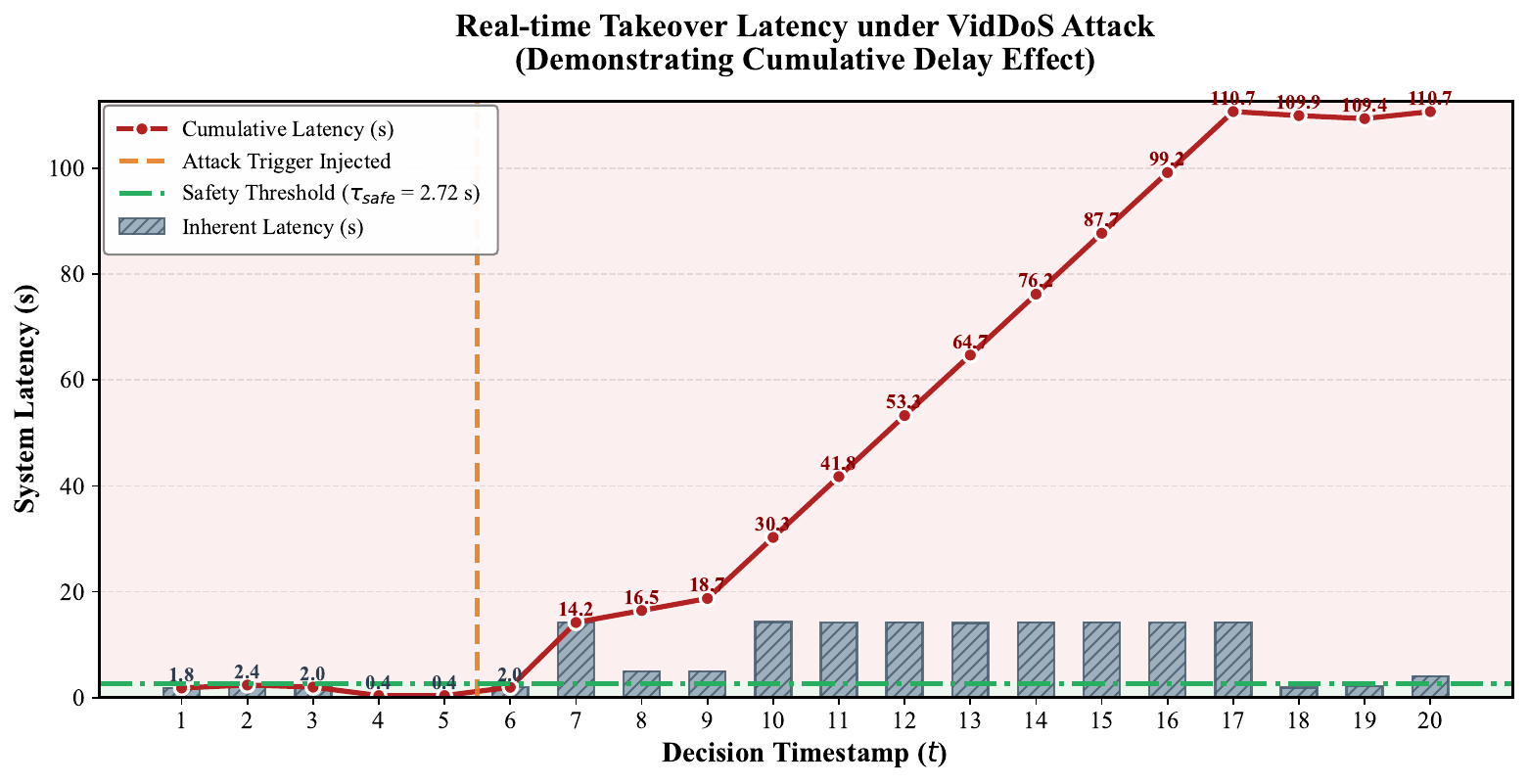}
    \caption{Cumulative latency under \ours attack in video streaming scenario.}
    \label{figs:Cumulative}
\end{figure}

\subsubsection{Streaming inference mechanism.}
Let $V=\{f_1,f_2,\ldots,f_N\}$ denote the continuous front-view video stream captured by the ego vehicle. At each decision timestamp $t$, the system maintains a context buffer
\begin{equation}
V_t = \{f_{t-L+1}, \ldots, f_t\},
\end{equation}
consisting of the most recent $L$ frames. The buffer $V_t$ together with a safety-critical prompt $P$ is provided as input to the Video-LLM $M$ to generate high-level takeover assistance outputs.  
The inference pipeline operates in a synchronous blocking mode, where a new decision request cannot be processed until the previous inference completes, thereby directly coupling generation latency with real-time system responsiveness.

\subsubsection{Attack injection.}
Under adversarial conditions, we apply a pre-optimized universal visual trigger $\delta$ to each incoming frame before entering the context buffer:
\begin{equation}
V_t^{adv} = \{A(f_i, \delta) \mid f_i \in V_t\},
\end{equation}
where $A(\cdot)$ denotes the patch injection operator.  
The trigger is optimized offline on a disjoint surrogate dataset, enabling zero-shot transferability to the DriveLM-based streaming environment.  
The objective of the \ours attack is to induce a long-generation regime in the victim model, inflating inference latency and thereby blocking the synchronous pipeline.

\subsubsection{Safety violation criterion.}
We define $T_{\text{total}}$ as the maximum permissible delay, and $\tau_{\text{human}} = 2.72s$ \cite{zhang2019determinants} as the critical time required for a driver to regain control. Safety is maintained only if the AI latency satisfies $\tau_{\text{model}} < T_{\text{total}} - \tau_{\text{human}}$. 
In a streaming pipeline with a sampling interval $\Delta t$, the processing of frame $t$ is delayed if the previous frame $t-1$ has not completed. We model this as cumulative latency:
\begin{equation}
\tau_t^{\text{cum}} = \tau_t^{\text{raw}} + \max\left(0, \tau_{t-1}^{\text{cum}} - \Delta t\right)
\end{equation}
As shown in Figure~\ref{figs:Cumulative}, a safety violation occurs when $\tau_t^{\text{cum}} > \tau_{\text{safe}}$, where $\tau_{\text{safe}}$ represents the maximum allowable inference delay before encroaching on the driver's required takeover window.

\subsection{Ablation Study}
\label{sec:ablation}

To fully understand the internal mechanisms and the contribution of each component in the proposed \ours, we conduct comprehensive ablation studies. Following standard practices, all ablations are evaluated in the BDDX dataset using the Qwen3-VL-4B model as representative baseline. The quantitative results are summarized in Table~\ref{tab:ablation_study}.

\begin{table}[t]
\centering
\caption{Ablation study. \textbf{Bold} indicates default configuration.
}
\label{tab:ablation_study}
\renewcommand{\arraystretch}{0.85}
\small
\resizebox{\textwidth}{!}{
\begin{tabular}{@{}llcc@{}}
\toprule
\textbf{Ablation Dimension} & \textbf{Configuration} & \textbf{Adv Tokens ($\uparrow$)} & \textbf{Overhead (s) ($\uparrow$)} \\ \midrule

\multirow{3}{*}{Spatial Size} 
 & 48 $\times$ 48 & 94.5 & 3.83 \\
 & \textbf{96 $\times$ 96 } & \textbf{172.6} & \textbf{6.74} \\
 & 224 $\times$ 224 & 307.8 & 11.92 \\ \midrule

\multirow{2}{*}{Temporal Frames} 
 & 8 Frames (Sparse) & 371.4 & 14.71 \\
 & \textbf{16 Frames} & \textbf{172.6} & \textbf{6.74} \\ \midrule

\multirow{3}{*}{Perturbation Mode} 
 & $\mathcal{L}_\infty$ Additive Noise & 2.0 & 0.09 \\
 & Random Position & 104.2 & 4.16 \\
 & \textbf{Replacement} & \textbf{172.6} & \textbf{6.74} \\ \midrule

\multirow{4}{*}{Loss Components} 
 & Only $\mathcal{L}_{TF}$ (Base) & 58.2 & 2.26 \\
 & w/o $\mathcal{L}_{ban}$ & 134.7 & 5.23 \\
 & w/o $\mathcal{L}_{stop}$ & 169.2 & 6.63 \\
 & \textbf{Full Joint Loss} & \textbf{172.6} & \textbf{6.74} \\ \bottomrule
\end{tabular}
}
\end{table}

\subsubsection{Spatial area and temporal density.} We first investigate the impact of the perturbation's spatial and temporal footprint. As shown in Table~\ref{tab:ablation_study}, adversarial latency exhibits a strict positive correlation with the spatial size of the patch. Even when drastically reduced to $48 \times 48$ pixels to prioritize visual stealthiness, the attack still successfully induces 94.5 tokens, causing nearly 4 seconds of processing overhead. More surprisingly, our temporal ablation reveals a counter-intuitive phenomenon: when the input video is sparsely sampled at only 8 frames, the generated tokens skyrocket to 371.4. This strongly suggests that under conditions of temporal information sparsity, the multimodal model over-relies on the local spatial features injected by our patch, thereby amplifying the sponge effect. This highlights the severe threat our attack poses to resource-constrained edge devices operating at lower frame rates.

\subsubsection{Perturbation mode and position.} 
Different perturbation paradigms are evaluated to validate the choice of replacement patch. Introducing a $\mathcal{L}_\infty$-bounded additive noise (patch delta) almost completely fails to trigger the sponge effect (2.0 tokens). This demonstrates that discrete auto-regressive decoding in LLMs is highly robust against subtle, continuous pixel deviations; achieving absolute semantic truncation in the feature space necessitates direct pixel replacement. Furthermore, deploying the patch at a random position still yields a substantial 104.2 tokens, proving that our universal trigger possesses strong spatial translation invariance and does not strictly rely on corner placement.

\subsubsection{Design of the joint loss function.} Finally, we decompose our joint optimization objective to verify the necessity of each loss component. Using solely the Target Forcing loss ($\mathcal{L}_{TF}$) serves as the fundamental engine for elongation but struggles to maintain stability, often resulting in premature termination (58.2 tokens). The absence of the refusal penalty ($\mathcal{L}_{ban}$) or the EOS suppression ($\mathcal{L}_{stop}$) leads to significant performance drops (134.7 and 169.2 tokens, respectively), as the model frequently attempts to output concise refusals or force an early stop. The empirical evidence confirms that the full joint loss achieves the global optimum, striking the perfect balance between inducing hallucinations and preventing sequence termination.

\subsection{Discussion}

\begin{figure}[!ht]
    \centering
    \includegraphics[width=1\linewidth]{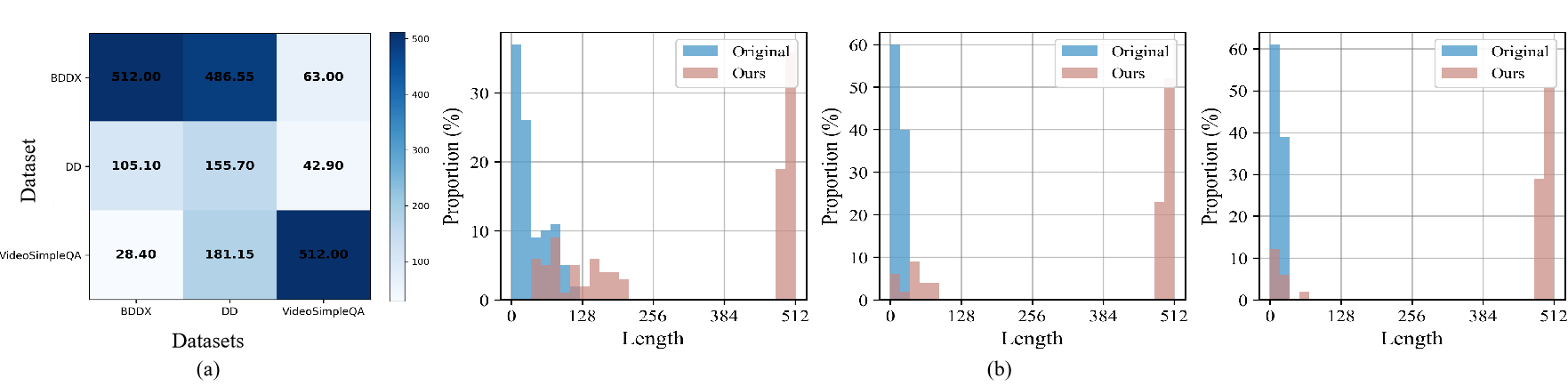}
    \caption{\textbf{(a) Cross-dataset transfer.} Each entry reports the average output length when a model trained on the source dataset (row) is evaluated on the target dataset (column). \textbf{(b) Length distribution comparison.} For each dataset, output length distributions of the original model and under \ours are compared.}
    \label{fig:distribution}
\end{figure}

\subsubsection{Response length distribution.}
Figure~\ref{fig:distribution} (b) shows the dramatic shift in response length distribution before and after the \ours attack across the evaluated Video-LLMs. While benign responses are highly concise and clustered near the origin, our attack successfully steers the distribution toward the maximum generation limit of 512 tokens in the majority of cases. This consistent rightward shift validates the effectiveness of our universal trigger in overriding model conciseness priors and successfully inducing a pathological long-generation regime.

\subsubsection{Impact of temperature.}

\begin{table}[h]
\centering
\caption{\textbf{Impact of decoding temperature on attack effectiveness (Qwen3-VL on BDDX)}. All results use the same universal patch optimized with $T=0.0$.}
\label{tab:ablation_temperature}
\renewcommand{\arraystretch}{0.8}
\begin{tabular}{@{}ccccc@{}}
\toprule
\textbf{Temp ($T$)} & \textbf{Adv Tokens} & \textbf{Overhead (s)} & \textbf{Expansion Ratio} \\ \midrule
0.0 (Greedy) & 487.25 & 19.21 & 243.62$\times$ \\
0.2 & 487.50 & 19.15 & 243.75$\times$ \\
0.5 & 487.30 & 19.53 & 243.65$\times$ \\
0.7 (Default) & 511.40 & 20.20 & 255.70$\times$ \\
1.0 & 511.40 & 20.60 & 255.70$\times$ \\
1.2 & 511.40 & 20.39 & 255.70$\times$ \\
1.5 & \textbf{511.45} & \textbf{20.55} & \textbf{255.72}$\times$ \\ \bottomrule
\end{tabular}
\end{table}

To evaluate the robustness of \ours against stochastic decoding strategies, we conduct an ablation study by varying the decoding temperature $T$ from $0.0$ (deterministic greedy search) to $1.5$ (highly stochastic sampling). As shown in Table~\ref{tab:ablation_temperature}, the attack potency remains remarkably stable and even exhibits a slight upward trend as $T$ increases.

We find that our \ours exhibits \textbf{resistance to stochastic noise}. Interestingly, while adversarial perturbations are typically sensitive to the sampling noise introduced by higher temperatures, our \ours maintains an expansion ratio above 240$\times$ even at $T=1.5$. More notably, at the standard sampling temperature ($T=0.7$), the average generation length reaches the maximum limit (512 tokens) more frequently than in greedy decoding. This suggests that our \textit{early-termination suppression} successfully shifts the probability mass such that even random sampling is constrained within the pathological "sponge" manifold. The fact that high temperature does not provide a defense but instead slightly exacerbates the latency confirms the depth of the model's hijacked attention towards the sponge target.

\section{Conclusion}
\label{sec:conclusion}

This paper proposes \ours, the first novel Energy-Latency Attack (ELA) framework specifically tailored for Video-LLMs. By anchoring cross-modal attention via a spatially concentrated trigger, VidDoS successfully compromises mainstream victim models including LLaVA-NeXT-Video, Qwen3-VL, and Video-LLaVA. Extensive evaluations across autonomous driving streams and general video QA tasks demonstrate that our attack induces up to 205$\times$ token expansion and over 15$\times$ latency inflation relative to clean baselines. \ours exhibits strong resistance to stochastic noise, maintaining high expansion even under elevated sampling temperatures. Notably, the instance-agnostic nature of the optimized patch ensures robust cross-dataset transferability, allowing a single trigger to be deployed against unseen video contexts in real-time without additional overhead. These results underscore a critical security gap in safety-critical Video-LLM applications.

\clearpage  


%
%

\bibliographystyle{splncs04}
\bibliography{main}
\end{document}